\documentclass[runningheads]{llncs}
\usepackage{eccv}
\usepackage{eccvabbrv}
\usepackage{graphicx}
\usepackage{booktabs}
\usepackage{multirow}
\usepackage[accsupp]{axessibility}  
\usepackage{comment}
\usepackage{hyperref}
\usepackage{orcidlink}

\begin{document}
\title{BelHouse3D: A Benchmark Dataset for Assessing Occlusion Robustness in 3D Point Cloud Semantic Segmentation} 
\titlerunning{BelHouse3D}
\author{Umamaheswaran Raman Kumar\inst{1}\orcidlink{0000-0001-7151-5050} \and
Abdur Fayjie\inst{1}\orcidlink{0000-0002-4139-4689} \and
Jurgen Hannaert\inst{2}\orcidlink{0009-0001-1495-6789} \and
Patrick Vandewalle\inst{1}\orcidlink{0000-0002-7106-8024}}
\authorrunning{U.~Raman Kumar et al.}
\institute{Dept. of Electrical Engineering (ESAT), KU Leuven, Belgium \and 3Frog, Belgium}

\maketitle

\begin{abstract}
Large-scale 2D datasets have been instrumental in advancing machine learning; however, progress in 3D vision tasks has been relatively slow. This disparity is largely due to the limited availability of 3D benchmarking datasets. In particular, creating real-world point cloud datasets for indoor scene semantic segmentation presents considerable challenges, including data collection within confined spaces and the costly, often inaccurate process of per-point labeling to generate ground truths. While synthetic datasets address some of these challenges, they often fail to replicate real-world conditions, particularly the occlusions that occur in point clouds collected from real environments. Existing 3D benchmarking datasets typically evaluate deep learning models under the assumption that training and test data are independently and identically distributed (IID), which affects the models' usability for real-world point cloud segmentation. To address these challenges, we introduce the BelHouse3D dataset, a new synthetic point cloud dataset designed for 3D indoor scene semantic segmentation. This dataset is constructed using real-world references from 32 houses in Belgium, ensuring that the synthetic data closely aligns with real-world conditions. Additionally, we include a test set with data occlusion to simulate out-of-distribution (OOD) scenarios, reflecting the occlusions commonly encountered in real-world point clouds. We evaluate popular point-based semantic segmentation methods using our OOD setting and present a benchmark. We believe that BelHouse3D and its OOD setting will advance research in 3D point cloud semantic segmentation for indoor scenes, providing valuable insights for the development of more generalizable models.

\end{abstract}

\begin{figure}[ht]
  \centering
  \includegraphics[page=1, width=\textwidth, trim = 0cm 19.1cm 0cm 0cm, clip]{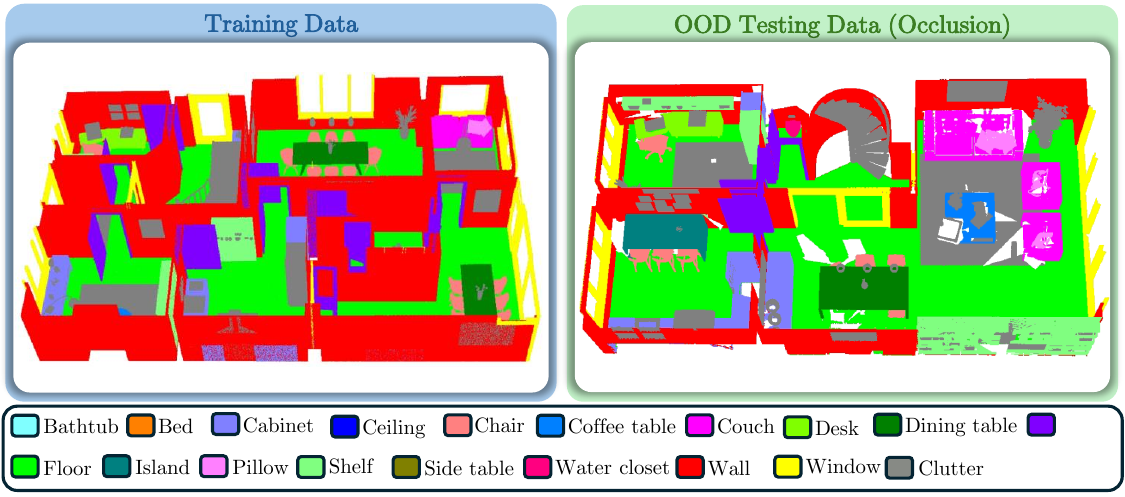}
  \caption{BelHouse3D is a synthetic 3D point cloud dataset specifically designed for indoor scene semantic segmentation. It provides clean data and precise annotations for model training and in-distribution (IID) testing. Additionally, it includes a test set designed to simulate real-world occlusion, serving as an out-of-distribution (OOD) benchmark. The left side of the figure displays a scene from the training dataset, representing House 6, while the right side illustrates the OOD test data with occlusion from House 30.}
  \label{fig:belhouse_intro}
\end{figure}

\section{Introduction}
In recent years, machine learning (ML) has achieved human-level performance (even beyond) in various computer vision tasks, such as image classification, segmentation, and object detection. These advancements have been largely driven by the availability of large-scale annotated 2D benchmarking datasets, such as ImageNet~\cite{deng2009imagenet}, TinyImageNet~\cite{le2015tiny}, PascalVOC~\cite{everingham2010pascal}, Omniglot~\cite{lake2019omniglot}, CIFAR-10~\cite{krizhevsky2009learning}, CUB-200~\cite{wah2011caltech}, MNIST and their derived datasets~\cite{lecun1998gradient, xiao2017fashion, cohen2017emnist} and many other~\cite{liu2015deep, nilsback2008automated, zhou2017places, young2014image, martin2001database, peng2019moment, fei2004learning, agustsson2017ntire, ros2016synthia}. However, the development of 3D understanding in machine learning remains challenging due to the complexities of data collection and annotation. Unlike 2D data, which provides a compressed pixel-based representation of the 3D world, 3D data, particularly in the form of point clouds, captures the spatial configuration of objects and scenes in a 3D Cartesian coordinate system. This allows point clouds to convey the spatial arrangement of visual features, which is crucial for understanding the structure of objects and scenes, rather than relying on texture as in images. As stated by Jagadeesh and Gardner~\cite{jagadeesh2022texture}, humans are more sensitive to spatial configurations than to texture, making point clouds particularly effective for studying 3D scene understanding.

\begin{table}[t]
    \caption{A summary of widely used 3D benchmarking datasets for indoor scene segmentation (SS), categorized by modality: point clouds (PC) and RGB-D, and data source: real-world (R) and synthetic (S). Additionally, we include other benchmarking tasks for these datasets: object classification (OC), object detection (OD), pose estimation (PE), depth estimation (DE), instance segmentation (IS), and novel-view synthesis (NVS).}
    \label{tab:existing_3D_segmentation_datasets}
    \centering
    \begin{tabular}{@{}l@{\hskip 0.3in} @{}c@{\hskip 0.3in} @{}c@{\hskip 0.3in} @{}l@{}}
    \toprule
    Dataset & Modality & R/S  & Benchmarking tasks\\
    \midrule
     NYUv2~\cite{silberman2011indoor}           & RGB-D  & R  & SS \\
     SunRGB-D~\cite{song2015sun}                 & RGB-D  & R  & SS, OC, OD, PE \\
     SceneNN~\cite{hua2016scenenn}              & RGB-D & R   & SS \\
     S3DIS~\cite{armeni20163d}                  & PC     & R  & SS       \\
     Matterport3D~\cite{chang2017matterport3d}  & RGB-D & R    & SS, DE \\
     ScanNet~\cite{dai2017scannet}              & RGB-D  & R  & SS, OC   \\
     PartNet (objects only)~\cite{mo2019partnet}& PC     & S  & SS, IS \\
     Hypersim~\cite{roberts2021hypersim}        & RGB-D  & S  & SS \\
     ARKitScenes~\cite{dehghan2021arkitscenes}  & RGB-D & R   & SS \\
     ScanNet200~\cite{rozenberszki2022language} & RGB-D  & S  & SS, OC   \\
     ScanNet++~\cite{yeshwanth2023scannet++}    & PC     & R  & SS, NVS  \\
     \midrule
     \textbf{BelHouse3D (Ours)} & \textbf{PC} & \textbf{S} &  \textbf{SS, FSS, OOD}\\
    \bottomrule
    \end{tabular}
\end{table}

Numerous 3D benchmarking datasets have been developed to facilitate the design and evaluation of machine learning models across various tasks, including classification~\cite{chang2015shapenet, wu20153d, roynard2018paris}, object detection~\cite{caesar2020nuscenes, sun2020scalability, ozsoy20224d, matuszka2022aimotive, song2015sun}, scene completion~\cite{caesar2020nuscenes, behley2019semantickitti, liao2022kitti, li2023sscbench}, and outdoor scene segmentation~\cite{roynard2018paris, xiao2022transfer, varney2020dales, hu2021towards, chen2022stpls3d, pan2020semanticposs, hackel2017semantic3d}. This work focuses on the segmentation of indoor scenes represented in point clouds. Based on their source of distribution, 3D datasets are categorized into real-world and synthetic datasets. \Cref{tab:existing_3D_segmentation_datasets} provides a summary of existing 3D datasets used for benchmarking on indoor scene semantic segmentation.  From the table, it can be observed that most real-world datasets utilize the RGB-D modality, which is often converted into point clouds. This preference for RGB-D is likely due to the relative ease of pixel-wise annotation in 2D, which can then be projected into 3D using depth information. Direct annotation of points in 3D is cumbersome, time-consuming, often inaccurate, and costly. Synthetic datasets address these challenges by simulating indoor environments using \emph{prelabeled} 3D assets and CAD models. Points are then sampled from the surfaces of these objects to construct the entire scene.

Fairness in synthetic data has been a significant concern within the research community, as biased data can lead to inaccurate predictions in downstream machine learning tasks. However, the notion of fairness in data is difficult to define. Van Breugel \etal emphasize the importance of fair synthetic data and advocate for causally-aware synthetic data generation~\cite{van2021decaf} . Approaches like VAEs~\cite{kingma2013auto} and GANs~\cite{xie2018differentially, goodfellow2020generative, yoon2020anonymization, gulrajani2017improved} have commonly been used to generate synthetic data through neural networks. However, this work adopts a manual 3D modeling and sampling technique for dataset creation to achieve more accuracy in labeling ground truths. To ensure fairness in the BelHouse3D  dataset, the synthetic data generation process is guided by real-world scene references, detailed in \cref{sec:dataset}.

Existing 3D benchmarking datasets for indoor scenes evaluate deep learning models under the implicit assumption that training and test data are independently and identically distributed (IID) from the same distribution. However, in real-world point cloud data, occlusion is inevitable, leading to changes in the shape, structure, and intrinsic geometry of a scene. Under such variations, characterized by out-of-distribution (OOD) sampling, a machine learning model is likely to degrade in performance and fail to provide reliable predictions. Evaluating models' performance under OOD conditions not only aids in understanding their generalization capabilities but also facilitates the design of more robust and trustworthy models. Despite its significance, progress in this area, particularly for \emph{3D indoor scene segmentation with point clouds}, is limited by the lack of a dedicated OOD benchmarking dataset.

\subsubsection{Contribution.} This paper introduces BelHouse3D, a novel 3D synthetic point cloud dataset designed for indoor semantic segmentation, capturing the geometry and structure of diverse household objects. Unlike existing synthetic datasets, BelHouse3D is developed using real-world references from 32 Belgian houses, ensuring greater fairness and reliability. The dataset provides a test set and configurations designed for OOD evaluation. Additionally, the paper presents a new benchmark for evaluating out-of-distribution (OOD) generalization in point-based fully-supervised and few-shot learning (FSL) segmentation models.

\subsubsection{Novelty.} The effect of out-of-domain (OOD) generalization, especially in the context of occlusion, a common occurrence in real-world point clouds, remains a relatively underexplored area. Although OOD generalization has been studied across various modalities, to the best of our knowledge, this study is the first to investigate OOD generalization in point cloud segmentation for 3D indoor scenes and to establish a dedicated benchmark for this task. 

\section{Related Work}
\subsubsection{Semantic segmentation.}
Existing 3D semantic segmentation methods are categorized into projection-based, discretization-based, hybrid, and point-based approaches~\cite{guo2020deep}. Projection-based methods project a 3D point cloud into 2D images, such as multi-view and spherical images. Discretization methods transform point clouds into intermediate volumetric or sparse representations. Hybrid methods leverage multi-modal features; for example, 3DMV~\cite{dai20183dmv} leverages a joint 3D-multiview network that combines RGB and geometric features. Chiang \etal introduced a unified point-based framework to learn 2D textual representation, 3D structures, and global features~\cite{chiang2019unified}, while MVPNet~\cite{jaritz2019multi} combines 2D multi-view images with spatial geometric features. 

Point-based methods, inspired by the pioneering work PointNet~\cite{qi2017pointnet}, directly input orderless and unstructured point clouds into a network to learn per-point features using shared multilayer perceptrons (MLPs) and pooling functions. Following PointNet, many point-based methods~\cite{qi2017pointnet++, jiang2018pointsift, zhao2019pointweb, hu2020randla, thomas2019kpconv, wang2019dynamic, ye20183d, zhao2019dar} have been developed, including point-wise MLP, point convolution, graph-based, and RNN-based methods. Point-wise MLP methods are the simplest, applying shared MLPs to achieve high-efficiency networks. Point convolution methods design convolutional operators for point clouds, enhancing the receptive field of point-wise MLP networks. Graph-based methods use graph neural networks to capture the underlying shape and geometry of 3D objects, offering higher accuracy but at the cost of increased computational expense. RNN-based methods employ Recurrent Neural Networks, which tend to be computationally intensive.

\subsubsection{3D indoor scene representation datasets.} Real-world datasets, NYUv2~\cite{silberman2011indoor}, SunRGB-D~\cite{song2015sun}, Sun3D~\cite{xiao2013sun3d}, and Matterport3D~\cite{chang2017matterport3d} datasets were designed by collecting low-resolution short RGB-D sequences. S3DIS~\cite{armeni20163d} dataset comprises 215 million points extracted from 272 indoor scenes, labeled with coordinate values, RGB color values, and surface normals for 13 distinct object classes. SceneNN~\cite{hua2016scenenn}, a 3D triangular mesh dataset includes data gathered from 100 indoor scenes, semantically annotated for 22 object classes. The ScanNet~\cite{dai2017scannet} dataset, which features labeled voxels rather than individual points or objects, comprises 1513 scans from 707 indoor environments, providing ground truth labels for 21 object categories. A key limitation of these datasets is their low geometric resolution, which hampers labeling accuracy. The recent ScanNet++~\cite{yeshwanth2023scannet++} dataset addresses this issue by incorporating high-quality geometry alongside color information from 460 indoor scenes to improve geometric fidelity. However, the scalability of ScanNet++ is limited by its data acquisition process, and its multi-label annotation scheme occasionally leads to labeling ambiguities.

Synthetic datasets such as SceneNet~\cite{hua2016scenenn}, Robotrix~\cite{garcia2018robotrix}, and House3D~\cite{wu2018house3d} were primarily designed to facilitate visual navigation tasks for artificial agents which offer 13, 38, and 80 semantically annotated object classes for segmentation, respectively. The InteriorNet~\cite{liInteriorNet18} dataset uses 158 distinct furniture models crafted by design experts to generate a comprehensive segmentation dataset. Jiang \etal~\cite{jiang2018configurable} introduce a configurable photorealistic dataset with 129 semantically annotated classes. The Hypersim dataset~\cite{roberts2021hypersim} generates 461 simulated indoor scenes using publicly available 3D assets, though it includes a limited number of annotated classes. The Structure3D dataset~\cite{zheng2020structured3d} utilizes 3.5K house designs, curated by professional designers, and is annotated with a `primitive + relationship' representation to capture the interrelations among objects comprehensively. However, these datasets are designed using simulated environments, which often fail to fully capture the complexity and variability of real-world conditions.

\subsubsection{OOD datasets.} Existing OOD datasets~\cite{kervadec2021roses, liang2023unknown, zhao2023ood, dong2024multiood, yu2020bdd100k, galil2023framework, hein2023or, wang2022vim, zhang2023nico++, nekrasov2024oodis, cho2024diagnostic} focus mostly on 2D image data and are commonly categorized into three types~\cite{zhao2022ood}: cross-domain, artificial corruptions, and synthetic data generation in controlled environments. Cross-domain approaches often use entirely different test datasets, which do not adequately address OOD shifts in object texture, shape, and context. Artificial corruptions, such as synthetic noise~\cite{hendrycks2019benchmarking} fail to represent real-world variations accurately. Similarly, synthetic data generation~\cite{kortylewski2018empirically}, while allowing precise control, may not fully capture the complexity of real-world scenarios. Kong \etal~\cite{kong2023robo3d} introduce Robo3D, which offers the SemanticKITTI-C, KITTI-C, WOD-C, and nuScenes-C test sets, designed from existing datasets for assessing OOD performance of 3D detection and segmentation models in outdoor environments, particularly for autonomous driving. This work addresses the lack of a 3D OOD benchmark point cloud segmentation for \emph{indoor scenes} by creating a synthetic point cloud dataset that includes data occlusions as OOD evaluation conditions.

\begin{figure}[t]
  \centering
  \includegraphics[page=1, width=\textwidth, trim = 1cm 18cm 1cm 1cm, clip]{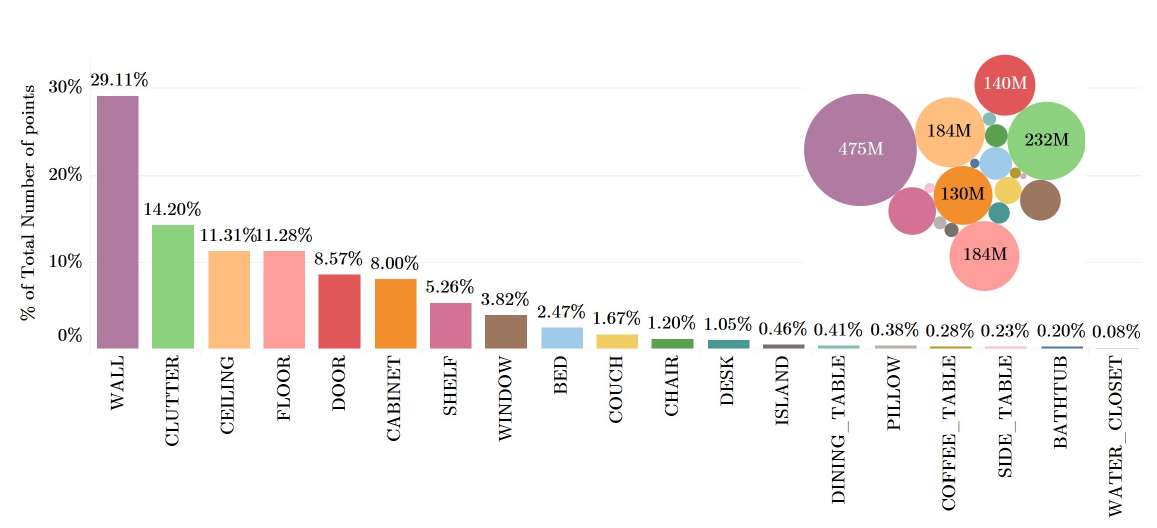}
  \caption{Bar graph illustrating the data distribution for the BelHouse3D dataset, with building structures comprising approximately $\approx64\%$, household objects $\approx22\%$, and clutter $\approx14\%$ of the total annotations. This distribution emphasizes the dataset's focus on capturing both the structural and functional elements of indoor scenes while addressing the challenge of segmenting clutter, a significant factor in real-world applications. Additionally, bubble sizes represent the number of points in each class, with the values (in millions) for the six major classes displayed.}
  \label{fig:data_distribution}
\end{figure}

\begin{figure}[t]
  \centering
  \includegraphics[page=1, width=\textwidth, trim = 0cm 0cm 0cm 15.3cm, clip]{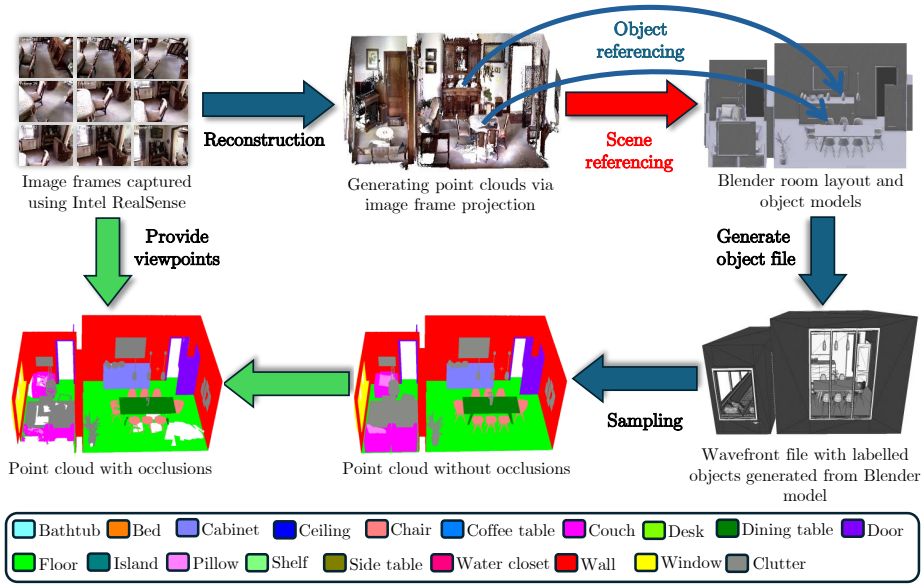}
  \caption{Workflow outlining the stages in creating the BelHouse3D dataset. The process starts with capturing image frames and proceeds through the following stages: point cloud reconstruction, creation of Blender files based on real data references, generation of object files, and final sampling to create the point clouds.}
  \label{fig:data_generation}
\end{figure}

\section{BelHouse3D Dataset} 
\label{sec:dataset}
The BelHouse3D dataset comprises 424 indoor scenes, collected from 32 distinct houses in Belgium offering accurate annotations for 19 classes. The classes are primarily categorized into three groups: 5 building structures (\emph{`ceiling'}, \emph{`floor'}, \emph{`wall'}, \emph{`door'} and \emph{`window'}), 13 household objects (\emph{`bathtub'}, \emph{`bed'}, \emph{`cabinet'}, \emph{`chair'}, \emph{`coffee table'}, \emph{`couch'}, \emph{`desk'}, \emph{`dining table'}, \emph{`island'}, \emph{`pillow'}, \emph{`shelf'}, \emph{`side table'}, and \emph{`water closet'}) and all other small objects collectively annotated as \emph{`clutter'}. The data distribution, illustrating the number of points for each object class, is shown in \cref{fig:data_distribution}. The subsequent section details the dataset generation process, covering the collection of real-world data, the creation of synthetic counterparts, and the OOD setting, as depicted in \cref{fig:data_generation}.

\subsection{Real-Data Acquisition}
\label{ss:real_data_acquisition}

The data acquisition process begins with capturing real-world data from 32 Belgian houses representing various indoor scenes. RGB-D images of a scene are obtained using an \emph{Intel RealSense} sensor and are then viewed in the \emph{Dot\textsl{3}D}~\cite{dotproduct} app running on an \emph{Apple ipad}. This setup facilitates systematic scanning of each room, even in confined spaces, by capturing multiple frames for comprehensive coverage. The frames, along with their corresponding camera poses, are saved as DotProduct~(.dp) files. Following the raw data collection and storage, a 3D reconstruction technique is employed to generate point clouds. This process involves projecting 2D pixels from each image frame, based on depth information, into 3D points defined in real-world coordinates (\textit{X}, \textit{Y}, and \textit{Z}). The 3D points from different frames are then registered to create a detailed 3D point cloud representation of the scanned environment.

\subsection{Synthetic Data Generation}
\label{sec:syn_data_gen}

Blender~\cite{blender}, a free and open-source 3D computer graphics software tool, is used to create the synthetic counterparts of the dataset. Real-world point clouds, detailed in \cref{ss:real_data_acquisition}, serve as references for constructing 3D models in Blender. Predefined object models are utilized for common household items to ensure cleaner data and precise annotations. The scenes, including these objects, are stored as wavefront files, from which point clouds are generated by sampling from the surfaces of the objects. To maintain fairness in the dataset, the data generation process is guided by real scene references through scene referencing and object referencing.

\subsubsection{Scene referencing.} In the scene referencing stage, room layouts are crafted in Blender to mirror the scale and dimensions of each room from real-world data. This reference aids in replicating the positions of building structures such as walls, ceilings, floors, doors, and windows from collected real-world data into the simulated environment. This approach ensures that the distribution of synthetic data closely aligns with that of its real-world counterpart.

\subsubsection{Object referencing.} Precise control in synthetic data generation is essential for achieving accurate data labeling, especially in 3D point clouds, where per-point labeling for semantic segmentation is labor-intensive, costly, and prone to inaccuracies. Object referencing offers greater control over data and label generation by using predefined object models to represent common household objects. These models are placed within the simulated room layouts based on the estimated positions of their real-world counterparts. However, this step does not strictly adhere to exact object positioning, allowing for controlled variations within object categories. Additionally, this method helps in filling occluded parts, resulting in a cleaner and more complete synthetic dataset.

\begin{figure}[t]
  \centering
  \includegraphics[page=2, width=\textwidth, trim = 0cm 15cm 0cm 0cm, clip]{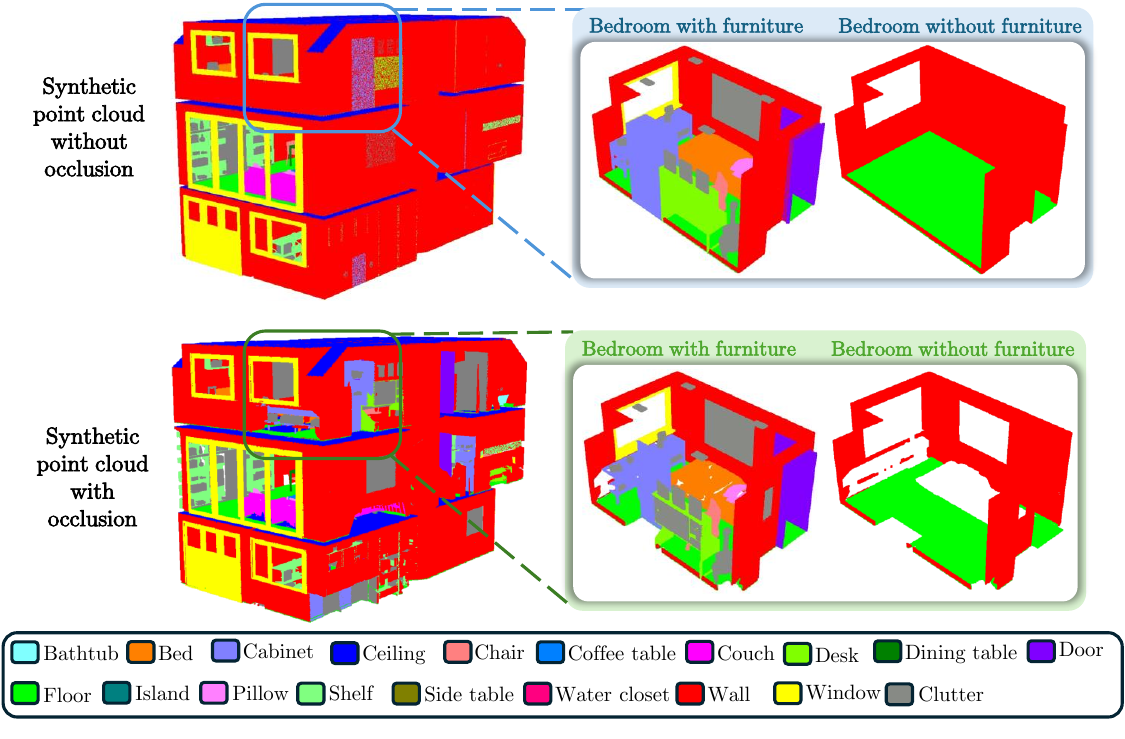}
      \caption{Synthetic point clouds generated for House 28 from the BelHouse3D dataset, comparing the IID test data (top) and the OOD test data with occlusion (bottom). The bedroom scene is depicted with and without furniture to highlight the impact of occlusion on building structures.}
  \label{fig:ood_setting}
\end{figure}

\subsection{OOD Setting}
To address the challenges posed by inevitable occlusions in real-world point clouds, a test set was meticulously developed by selectively removing points from specific portions of objects or scenes within a controlled simulated environment. This process involves using the original viewpoints from which the real-world data was acquired to filter the points of the synthetic point cloud generated in \cref{sec:syn_data_gen}, retaining only those visible in the selected viewpoints while discarding the rest (shown in \cref{fig:ood_setting}). This approach ensures that the occlusions are introduced in a manner that closely mimics real-world scenarios, rather than removing points arbitrarily, which could otherwise result in the unintended loss of entire objects. The OOD evaluation is crucial for assessing the models' generalization capabilities and their practical applicability in real-world point cloud segmentation.

\section{Benchmarking and Discussions}
\label{sec:experiments}
This section presents the benchmarking results for 3D semantic segmentation under fully supervised and few-shot settings, evaluated in both IID and OOD conditions. The standard evaluation metrics for 3D semantic segmentation, including mean Intersection over Union (mIOU) and Overall Accuracy (OA), are used for benchmarking.

\subsection{Fully Supervised 3D Segmentation in OOD Conditions}
\label{sec:sem_seg}
3D Semantic Segmentation of point clouds is a computer vision task that involves the precise labeling of individual points within a three-dimensional space. One of the primary challenges in 3D Semantic Segmentation arises from the irregular and unstructured nature of point clouds. Unlike traditional 2D images, point clouds lack a predefined grid structure, making it challenging to apply conventional image processing techniques directly. Point-based methods, directly operating on point clouds, have emerged as promising approaches to address these challenges.

\subsubsection{Dataset configuration.} For fully supervised methods, a \textsl{60-20-20} data split is implemented, dividing the dataset into three distinct subsets: the \textit{train} set, which includes data from House 1 to House 20; the \textit{validation} set, consisting of data from House 21 to House 26; and the \textit{test} set, comprising data from House 27 to House 32. Given the substantial size of the BelHouse3D dataset, preprocessing is performed in advance, with files organized into the respective train, validation, and test folders to streamline the training, validation, and evaluation processes.

During the training of non-transformer models, a random subset of 2048 points is selected from each point cloud block within the \textit{train} folder. Conversely, during the validation and testing phases, all available samples from the respective folders are employed to ensure a thorough evaluation of the model's performance. For training transformer models, Pointcept~\cite{pointcept2023} is extended, utilizing sparse tensor representation, thereby enabling evaluation for transformer models.

\begin{table}[!h]
  \caption{Benchmarking results for fully supervised 3D semantic segmentation, reported as \textit{mIoU} and \textit{OA}. Results are shown for both IID and OOD (occlusion) test sets.}
  \label{tab:semseg-benchmark}
  \centering
  \begin{tabular}{@{}r@{\hskip 0.1in}cc@{\hskip 0.1in}cc@{\hskip 0.1in}c@{\hskip 0.1in}c@{}}
    \toprule
     && \multicolumn{2}{c}{IID} && \multicolumn{2}{c}{OOD (occlusion)} \\
    \cmidrule{3-4} \cmidrule{6-7}
          && mIoU & OA && mIoU & OA \\
    \midrule
    Pointnet~\cite{qi2017pointnet}     && 38.38 & 66.16 && 26.97 ($-30\%$) & 68.89 ($+4\%$)\\
    Pointnet++ \cite{qi2017pointnet++} && 71.97 & 82.33 && 36.56 ($-49\%$) & 76.06 ($-8\%$)\\
    DGCNN \cite{wang2019dynamic}       && 72.57 & 84.85 && 40.34 ($-44\%$) & 76.92 ($-11\%$)\\
    Stratified Transformer \cite{lai2022stratified} && 79.12 & 89.12 && 44.54 ($-44\%$) & 78.24 ($-12\%$)\\
    Point TransformerV2 \cite{wu2022point} && 82.32 & 88.44 && 55.95 ($-32\%$) & 80.83 ($-9\%$)\\
    \bottomrule
  \end{tabular}
\end{table}

\begin{figure}[t]
  \centering
  \includegraphics[page=2, width=\textwidth, trim = 0cm 0cm 0cm 15cm, clip]{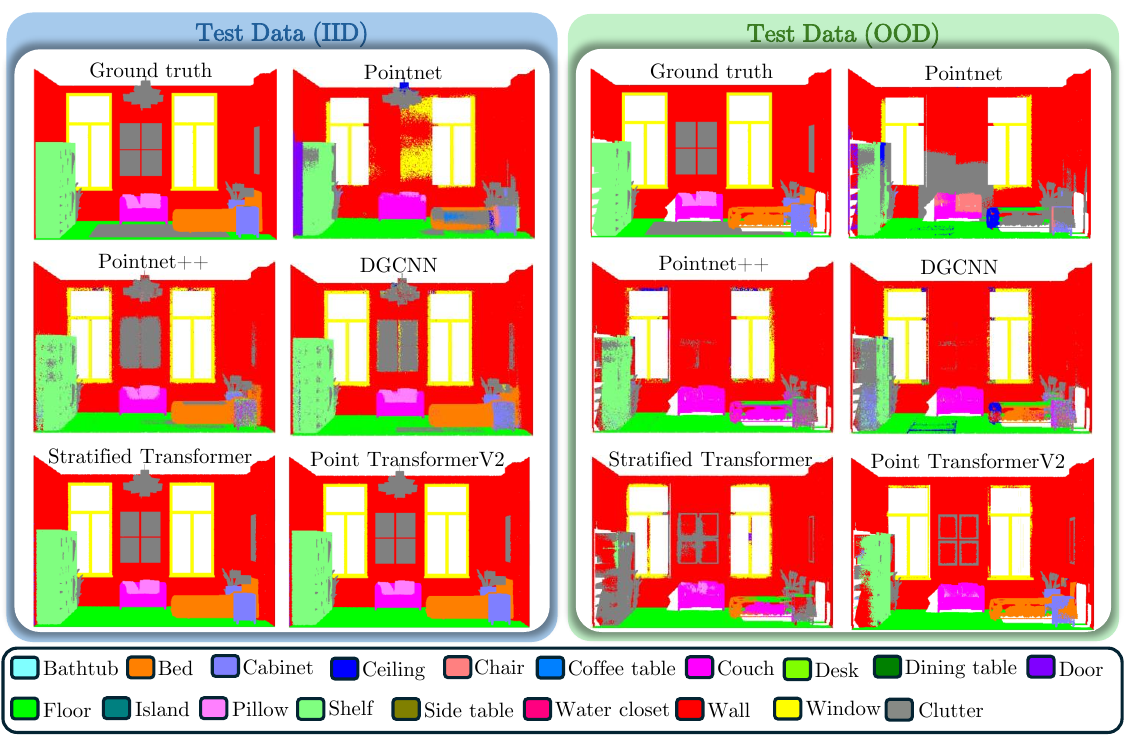}
  \caption{Qualitative results of fully supervised 3D segmentation methods evaluated on House 30 from the BelHouse3D dataset. The performance decline is evident across all methods when transitioning from the IID test set to the OOD test set with occlusions.}
  \label{fig:sem_seg_ood}
\end{figure}

\subsubsection{Benchmarking.}
The benchmarking results presented in ~\cref{tab:semseg-benchmark} reveal a significant decline in model performance when transitioning from IID to OOD (occlusion) test sets, as indicated by the decrease in mIoU across all models. For instance, models like PointNet++ and DGCNN show significant performance degradation, with mIoU reductions of 49\% and 44\%, respectively. The Stratified Transformer and Point TransformerV2, although performing better than earlier models in the IID scenario, still exhibit significant drops in mIoU (44\% and 32\%, respectively) under the OOD scenario. These results highlight the sensitivity of 3D semantic segmentation models to occlusion, a common challenge in real-world point cloud data.

Interestingly, while mIoU scores decrease under OOD conditions, OA remains relatively stable or even improves slightly in some cases, such as with PointNet, which sees a 4\% increase in OA. The discrepancy between mIoU and OA across the models suggests that occlusion predominantly impacts the segmentation accuracy of smaller classes, such as household objects, more compared to larger, more prominent classes like building structures. These findings underscore the importance of developing more robust models capable of generalizing to OOD conditions, particularly in the context of 3D indoor scene segmentation where occlusion is prevalent. Additionally, \cref{fig:sem_seg_ood} provides qualitative results of fully supervised 3D segmentation methods benchmarked on BelHouse3D.

\subsection{Few-Shot 3D Segmentation in OOD Conditions}
\label{sec:fss}

Few-shot 3D semantic segmentation represents a recent approach in computer vision, addressing the challenge of accurate segmentation with minimal labeled data. Unlike conventional segmentation tasks, which often require extensive training datasets, few-shot segmentation aims to generalize effectively from a limited number of annotated examples. This approach facilitates the rapid adaptation of models to new, previously unseen environments.

\subsubsection{Dataset configuration.} The data split for the few-shot setting deviates from its fully supervised counterpart due to the adoption of episodic training in few-shot scenarios. As few-shot testing assesses the performance of a network trained on \textit{Base} classes while evaluated on \textit{Novel} classes, the dataset is not separated into distinct train, validation, and test splits. Instead, it is partitioned based on class labels for training (\textit{Base}) and evaluation (\textit{Novel}) to ensure the absence of class overlaps. The \textit{Base} class split contains (\emph{`bathtub'}, \emph{`cabinet'}, \emph{`ceiling'}, \emph{`desk'}, \emph{`island'}, \emph{`pillow'}, \emph{`side table'}, \emph{`water closet'}, \emph{`window'}) and the \textit{Novel} class split contains (\emph{`bed'}, \emph{`chair'}, \emph{`coffee table'}, \emph{`couch'}, \emph{`dining table'}, \emph{`door'}, \emph{`floor'}, \emph{`shelf'}, \emph{`wall'}, \textit{`clutter'}).

In the FSL setting, a model is trained using \textit{N}-way \textit{K}-shot task. Each episode involves randomly selecting \textit{N} classes from the list of training classes. For each of the \textit{N} selected classes, \textit{K} support samples are randomly sampled from the point cloud. Unlike training, where the support set is randomly selected from all the samples during each episode, testing establishes a fixed test set. To achieve this, initially, $\textit{N} \times \textsl{5}$ samples are selected from the entire sample list for each class and designated as the support set, with the remaining samples forming the query set. This approach replicates a real-world scenario with considerably fewer support samples than query samples. For each \textit{N}-way \textit{K}-shot scenario, a separate test set folder is created. Each folder contains a maximum of \textsl{200} episodes per combination of \textit{N}-way classes without repetition in the query set for each episode, in contrast to the training case.

\begin{table}[t]
  \caption{Benchmarking results for few-shot 3D semantic segmentation, reported as \textit{mIoU} for both \textit{Base} and \textit{Novel} classes under IID and OOD (occlusion) test sets.}
  \label{tab:fss-benchmark}
  \centering
  \begin{tabular}{@{}r@{\hskip 0.1in}@{\hskip 0.1in}c@{\hskip 0.1in}cc@{\hskip 0.1in}c@{\hskip 0.1in}c@{}}
    \toprule
     Model & \multicolumn{2}{c}{IID} && \multicolumn{2}{c}{OOD (occlusion)} \\
    \cmidrule{2-3} \cmidrule{5-6}
       & Base & Novel && Base & Novel \\
    \midrule
    $1-way$ $1-shot$ \\
    ProtoNet~\cite{snell2017prototypical, zhao2021few} & 52.32 & 46.37 && 50.41 ($-3.65\%$) & 46.33 ($-0.09\%$)  \\
     AttProto~\cite{zhao2021few} & 51.68 & 46.93 && 50.26 ($-2.74\%$) & 44.61 ($-4.94\%$)  \\
     MPTI~\cite{zhao2021few} & 53.09 & 49.87 && 53.11 ($+0.04\%$) & 53.28 ($+6.84\%$)  \\
     AttMPTI~\cite{zhao2021few} & 49.45 & 47.17 && 51.21 ($+3.56\%$) & 51.07 ($+8.28\%$)  \\
    \midrule
    $1-way$ $5-shot$ \\
    ProtoNet~\cite{snell2017prototypical, zhao2021few} & 58.72 & 56.71 && 54.30 ($-7.53\%$) & 56.39 ($-0.56\%$)  \\
     AttProto~\cite{zhao2021few} & 59.91 & 59.46 && 56.94 ($-4.96\%$) & 59.24 ($-0.37\%$)  \\
     MPTI~\cite{zhao2021few} & 62.57 & 61.06 && 59.43 ($-5.02\%$) & 64.13 ($+5.03\%$)  \\
     AttMPTI~\cite{zhao2021few} & 62.69 & 61.75 && 58.98 ($-5.92\%$) & 64.84 ($+5.00\%$)  \\
    \midrule
    $2-way$ $1-shot$ \\
     ProtoNet~\cite{snell2017prototypical, zhao2021few} & 45.68 & 38.61 && 39.51 ($-13.51\%$) & 38.06 ($-1.42\%$)  \\
     AttProto~\cite{zhao2021few} & 44.95 & 38.28 && 40.24 ($-10.48\%$) & 38.20 ($-0.21\%$) \\
     MPTI~\cite{zhao2021few} & 38.54 & 40.59 && 37.26 ($-3.32\%$) & 44.06 ($+8.55\%$) \\
     AttMPTI~\cite{zhao2021few} & 34.59 &  38.61 && 30.72 ($-11.19\%$) & 39.49 ($+2.28\%$) \\
    \midrule
    $2-way$ $5-shot$ \\
     ProtoNet~\cite{snell2017prototypical, zhao2021few} & 50.95 & 45.31 && 47.14 ($-7.48\%$) & 45.49 ($+0.40\%$)  \\
     AttProto~\cite{zhao2021few} & 54.98 & 46.52 &&  47.26 ($-14.04\%$) & 45.49 ($-2.21\%$) \\
     MPTI~\cite{zhao2021few} & 47.92 & 48.88 &&  47.47 ($-0.94\%$) & 49.96 ($+2.21\%$) \\
     AttMPTI~\cite{zhao2021few} & 51.17 & 51.93 &&  50.67 ($-0.98\%$) & 52.03 ($+0.19\%$) \\
    \bottomrule
  \end{tabular}
\end{table}

\begin{figure}[ht]
  \centering
  \includegraphics[page=3, width=\textwidth, trim = 0cm 14cm 0cm 0cm, clip]{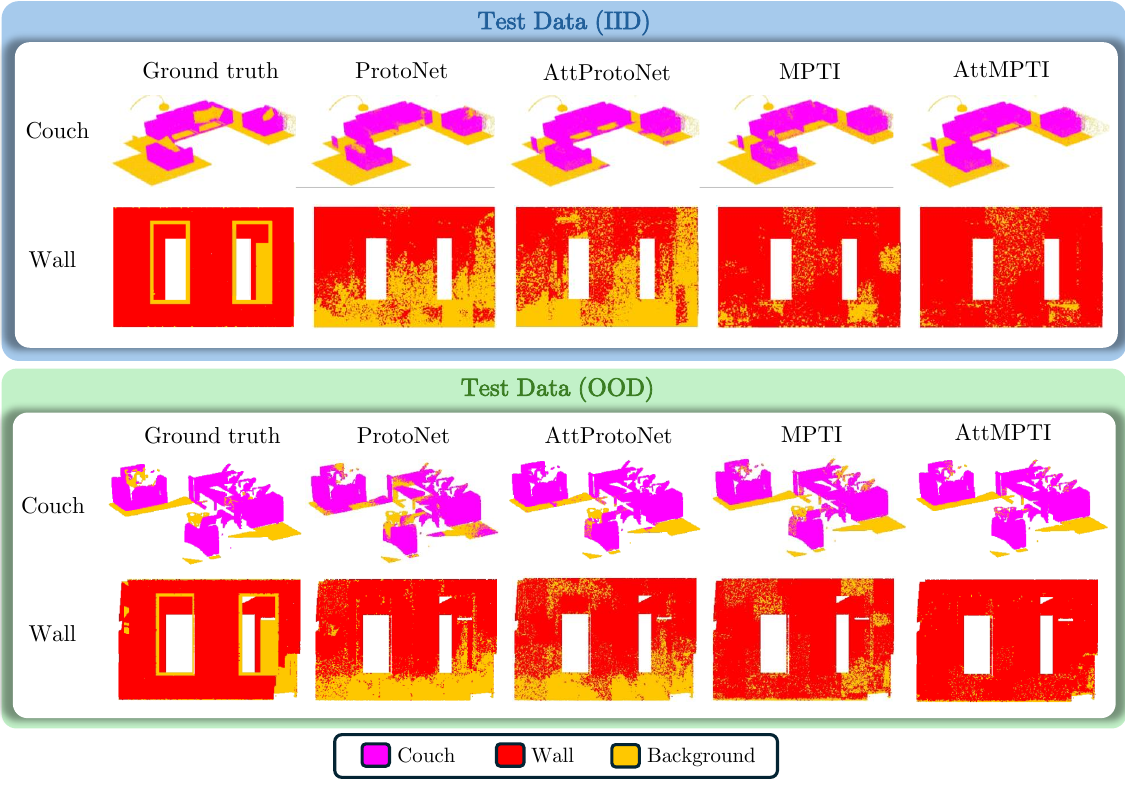}
  \caption{Qualitative results of few-shot 3D segmentation methods in the `1-way 5-shot' setting evaluated on House 30 from the BelHouse3D dataset. The performance across all methods shows minimal variation when transitioning from the IID test set to the OOD test set with occlusions.}
  \label{fig:fss_ood}
\end{figure}

\subsubsection{Benchmarking.}

The benchmarking results for few-shot 3D semantic segmentation, as presented in \cref{tab:fss-benchmark}, reveal distinct patterns in model performance across both IID and OOD (occlusion) test conditions. For the `1-way 1-shot’ and `1-way 5-shot’ settings, models like ProtoNet and AttProto generally show a slight decrease in mIoU when transitioning from IID to OOD settings, particularly for \textit{Base} classes. However, the performance for \textit{Novel} classes tends to remain stable or even improve slightly, as seen in the MPTI and AttMPTI models. This indicates that these methods may be better equipped to handle novel categories in occluded scenarios, suggesting that the integration of attention mechanisms or task-specific strategies could enhance model adaptability under challenging OOD conditions.

In contrast, the `2-way’ scenarios present a more mixed outcome. For instance, while the ProtoNet and AttProto models exhibit noticeable declines in mIoU for both \textit{Base} and \textit{Novel} classes under OOD conditions, models like MPTI and AttMPTI demonstrate relatively stable or improved performance, especially for \textit{Novel} classes. The MPTI and AttMPTI models, in particular, appear to benefit from their advanced feature integration techniques, which may help them better generalize in multi-class few-shot settings, even when faced with occlusions. The results indicate that few-shot learning methods exhibit better robustness in handling OOD challenges compared to fully supervised approaches. Additionally, \cref{fig:fss_ood} provides qualitative results of few-shot 3D segmentation methods benchmarked on BelHouse3D.

\section{Limitation And Future Work}
The BelHouse3D dataset significantly enriches the pool of available 3D datasets for indoor scene segmentation by providing clean data and accurate ground truth annotations. However, the current dataset is limited in the number of object classes and point cloud attributes ($XYZ$ only). Future work will focus on expanding the number of annotated classes, particularly for smaller household objects. Additionally, we aim to expand the dataset's scope to accommodate new tasks such as instance segmentation and scene completion, as well as more out-of-distribution (OOD) conditions, including variations in context (e.g., lighting, the connectedness of objects), object poses (e.g.\ pillow in various poses), and object shapes (e.g., different sofa models).

\section{Conclusion}
In this paper, the BelHouse3D dataset is introduced as a novel synthetic 3D dataset designed for indoor scene segmentation. The dataset contains realistic and clean point clouds that accurately reflect real-world houses, aiding deep learning models in understanding the underlying geometry of indoor environments. BelHouse3D also includes a test set specifically designed to evaluate model robustness under out-of-distribution (OOD) conditions, particularly addressing the challenge of data occlusion, which is common in real-world point cloud scenes. A benchmark is established through the evaluation of popular point-based fully supervised and few-shot segmentation methods within this OOD framework. Experimental results demonstrate that few-shot learning (FSL) can effectively address OOD challenges with minimal labeled data, indicating a promising direction for future segmentation models to enhance OOD robustness by incorporating 3D object geometry to better manage occlusions.

\bibliographystyle{splncs04}
\bibliography{ref}
\end{document}